\documentclass[journal,twoside,web]{ieeecolor}
\usepackage{xcolor}
\usepackage{generic}
\usepackage{cite}
\usepackage{amsmath,amssymb,amsfonts}
\usepackage{algorithmic}
\usepackage{graphicx}
\usepackage{textcomp}
\def\BibTeX{{\rm B\kern-.05em{\sc i\kern-.025em b}\kern-.08em
    T\kern-.1667em\lower.7ex\hbox{E}\kern-.125emX}}
\begin{document}
\title{Identifying contributors to supply chain outcomes in a multi-echelon setting: a decentralised approach}
\author{Stefan Schoepf, Jack Foster, and Alexandra Brintrup 
\thanks{Manuscript received 2 June 2023; revised 3 May 2024; accepted
20 July 2024. This work was supported by the Accenture Turing Strategic
Partnership, the Turing Enrichment scheme, EPSRC CDT AgriFoRwArdS [EP/S023917/1], and UKRI EPSRC DTP [EP/W524633/1].}
\thanks{Stefan Schoepf, Jack Foster, and Alexandra Brintrup are with the Supply Chain AI Lab of the Department of Engineering at the University of Cambridge, Cambridge, 17 Charles Babbage Rd, CB3 0FS UK and The Alan Turing Institute, British Library, 96 Euston Rd., London NW1 2DB (e-mail: \{ss2823, jwf40, ab702\}@cam.ac.uk)}
}

\maketitle

\begin{abstract} 
Organisations often struggle to identify the causes of change in metrics such as product quality and delivery duration. This task becomes increasingly challenging when the cause lies outside of company borders in multi-echelon supply chains that are only partially observable. Although traditional supply chain management has advocated for data sharing to gain better insights, this does not take place in practice due to data privacy concerns. We propose the use of explainable artificial intelligence for decentralised computing of estimated contributions to a metric of interest in a multi-stage production process. This approach mitigates the need to convince supply chain actors to share data, as all computations occur in a decentralised manner. Our method is empirically validated using data collected from a real multi-stage manufacturing process. The results demonstrate the effectiveness of our approach in detecting the source of quality variations compared to a centralised approach using Shapley additive explanations.

\end{abstract}


\section{Introduction}
\label{sec:introduction}

\begin{figure}[t!]
\centerline{\includegraphics[width=\columnwidth]{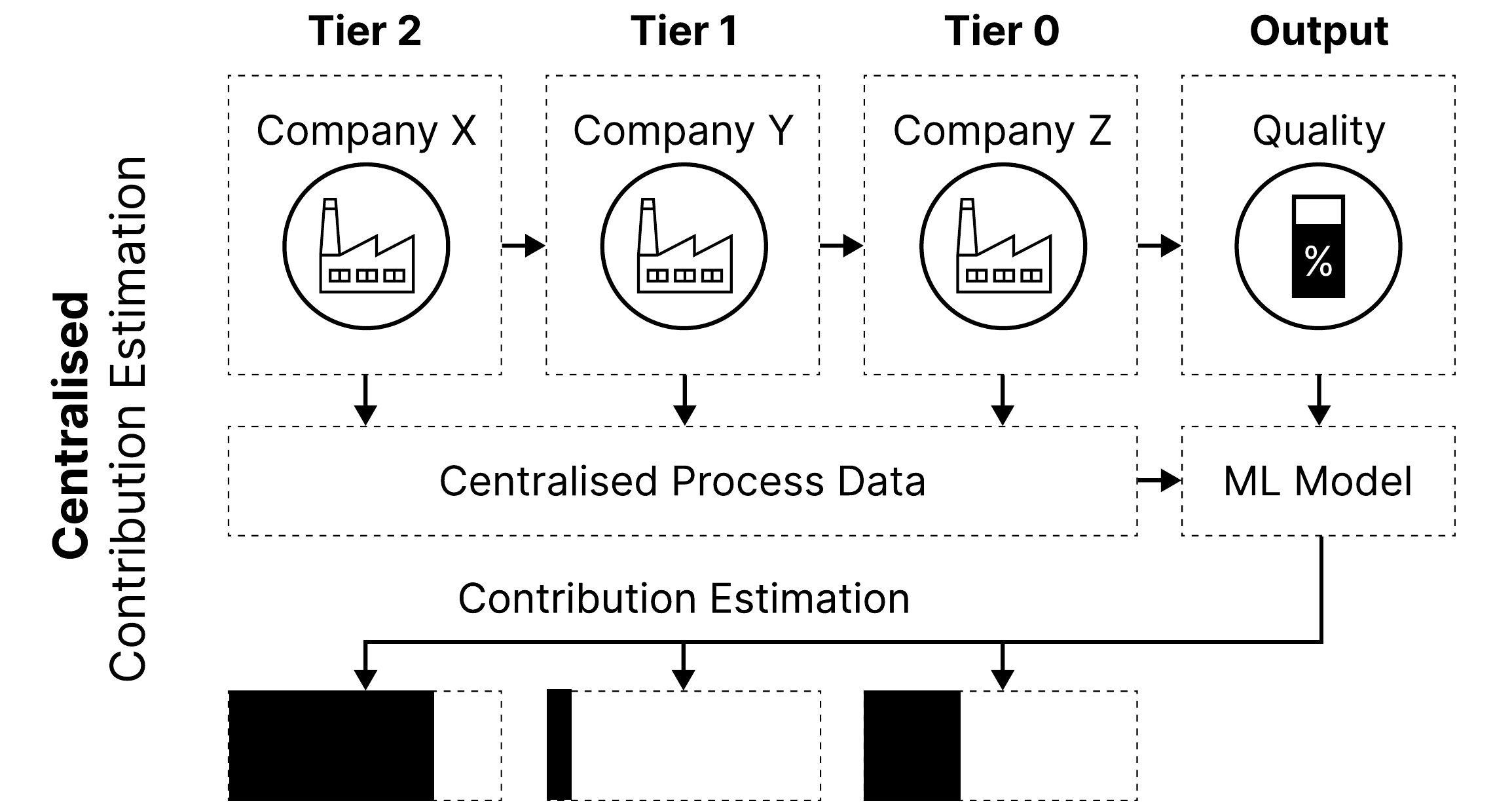}}
\caption{A centralised contribution estimation setting in which the machine learning (ML) model has access to data from multiple companies.}
\label{fig:central}
\end{figure}

\begin{figure}[t!]
\centerline{\includegraphics[width=\columnwidth]{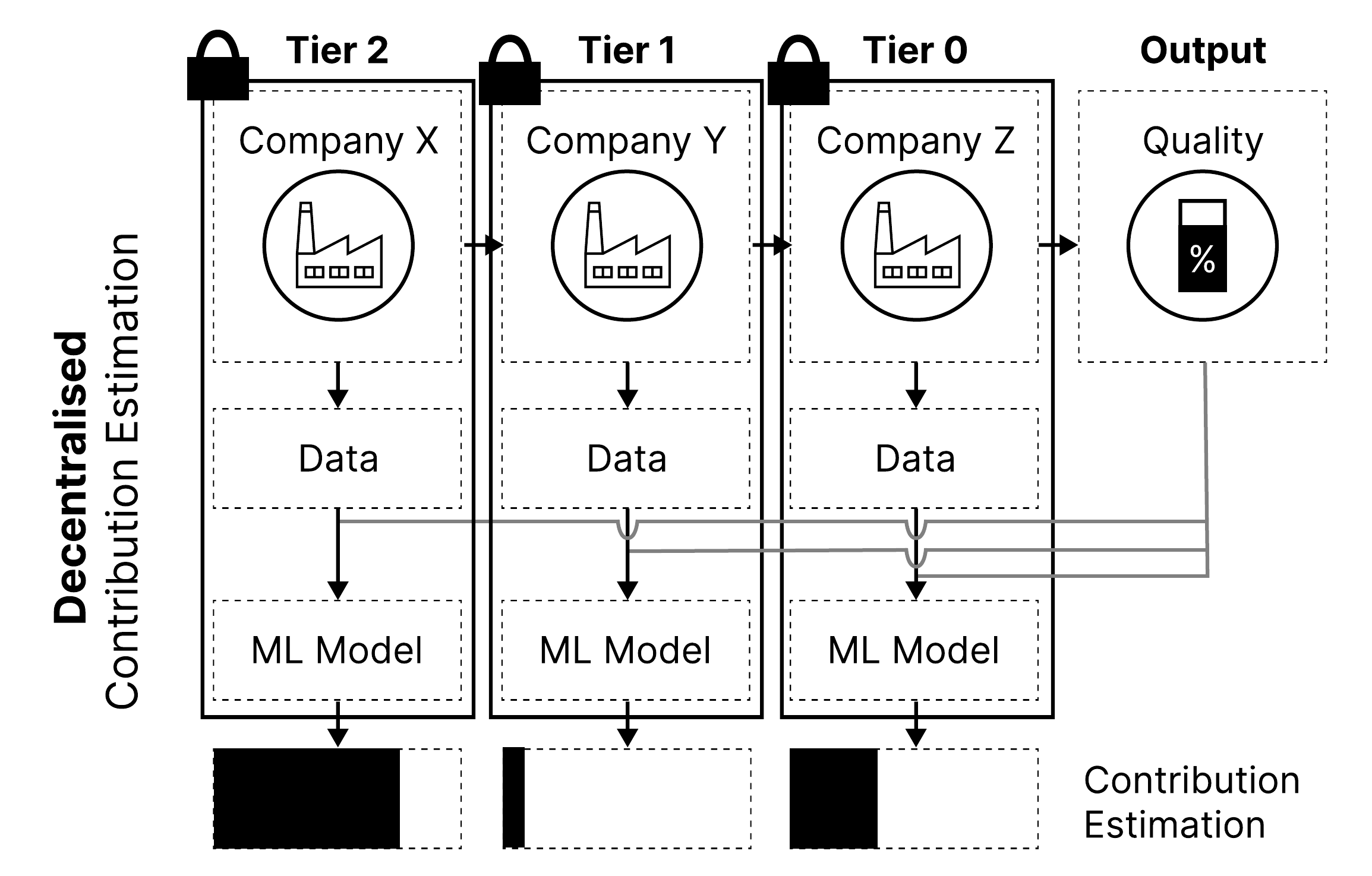}}
\caption{A decentralised contribution estimation setting in which the data of each company is not shared with others.}
\label{fig:decentral}
\end{figure}

\IEEEPARstart{T}{he} detection of causes for an emergent problem is an essential step in any process optimisation task. This is especially true for Supply Chain Management (SCM), an industry that requires reliable and efficient processes to meet customer demands in competitive low-margin settings. When a manufacturer struggles with quality issues that could be caused by any of its suppliers in a multi-echelon supply chain, it is non-trivial to uncover the source of the problem. Numerous methods exist in SCM and manufacturing research to estimate the influence (or contribution) of process parameters on metrics of interest that rely on a central data pool or other forms of data exchange between companies, as shown in Fig. \ref{fig:central} for a supply chain setting \cite{zantek2002process} \cite{senoner2022using} \cite{meister2021investigations} \cite{brito2022explainable}. 
We refer to contributors as processes or bundles of processes (e.g., a company) that together create an output and have an influence on the outcome of this output. Fig. \ref{fig:central} illustratively shows a process made up of three companies that produce a product and each has a different quantifiable contribution to the quality variation of this output.
Studies such as \cite{zantek2002process} \cite{senoner2022using} \cite{meister2021investigations} \cite{brito2022explainable} have shown promising potential in various industrial applications including the identification of the key processes that contribute to quality variations.
However, their applicability is limited to situations where it is feasible to train a model with centralised data from all process steps, which is not always feasible in real-life settings. This is particularly true for processes that are spread across multiple companies that are unwilling to share data due to privacy concerns, i.e. a typical supply chain, and would thus need a method without data sharing as shown in Fig. \ref{fig:decentral}. 

Due to this lack of process-level data sharing, companies are severely hindered in their ability to detect problem sources. The need for methods that work without violating data privacy can be seen in well-established works such as \cite{new_barbarossa2007decentralized}, where the authors use a decentralised maximum-likelihood estimation (MLE) process to synchronise dynamical systems. Their decentralized MLE approach builds upon the local exchange of the state functions. While this avoids the need to send data to a centralized fusion centre, as described in \cite{new_barbarossa2007decentralized}, it does not meet the privacy demands of our problem setting. Companies sharing their internal state with each other would pose an unacceptable data exposure. More recent literature reviews such as \cite{new_jiang2021differential} and studies show data aggregation \cite{new_shang2023robust} and consensus finding \cite{new_ying2023privacy} methods with data privacy for multi-agent systems but due to their design for data privacy, they are not applicable for the task of determining how much each individual participant contributed to an outcome. Their design to protect data privacy is at odds with the goal of inferring the influence of a participant's data on the overall outcome. On top of this problem, studies such as \cite{new_tableak_fl, new_FL1} have shown that systems that build models with data from multiple participants are prone to attacks that exploit information leakage from model weights/parameters. It is therefore understandable, that SC participants do not want their data to be shared, even when aggregation methods seem secure at a first glance.

The economic incentive for a method that can bridge this challenge of data privacy and insights is clear. 
The detection of a quality problem in a product whose components are sourced from multiple suppliers in a multi-echelon supply chain can cause a costly stream of audits to uncover the cause. Intangible problems such as variance in delivery delays can be even harder to investigate without the underlying data. If a supplier far upstream in the supply chain has a process that causes high variance multiple tiers further downstream in the supply chain, it is unlikely that this cause can be detected without access to the process-level data from multiple companies.
In order to overcome the problem of data privacy, the estimation of contribution to the metric of interest (e.g., quality, delay) needs to be performed without sharing the process-level data amongst supply chain participants. This problem setting, therefore, raises the research question: How can we estimate the contribution of companies to a supply chain outcome without sharing any of their internal data to ensure data privacy?
We propose a framework using Explainable Artificial Intelligence (XAI), more specifically neural network ensemble uncertainty \cite{lakshminarayanan2017simple}, to estimate contributions to a metric of interest in a decentralised manner. This ensures data privacy for the participating organisations, as only the estimated contribution is shared whereas data stays with the company for the local computation. We benchmark our method against the most popular XAI method in SCM literature, SHapley Additive exPlanations (SHAP) \cite{lundberg2017unified}. Both methods are applied to a multi-echelon process-level scenario. The SHAP approach uses all process data centralised for one model. Results show that our decentralised method achieves comparable results to SHAP while preserving data privacy due to its decentralised approach. We makes three main contributions to the literature:
\begin{enumerate}
    \item We introduce a decentralised approach for contribution estimation using neural network ensemble uncertainty, which acts as the first method to work under the strict no-data-sharing restrictions necessary for supply chains as shown in Fig. \ref{fig:decentral}. 
    \item We provide empirical validation of the proposed decentralised approach with real-life manufacturing data compared to a centralised SHAP approach. 
    \item We lay out further promising applications for decentralised contribution estimation (e.g., federated learning participant selection).
\end{enumerate}

The remainder of this work is structured as follows. In section \ref{sec:related} we review the literature and reveal shortcomings of existing works for the application in SCM. In section \ref{sec:method} we describe our method and the data used. In section \ref{sec:results} we benchmark our method against a centralised SHAP approach before concluding in section \ref{sec:conclusion}.

\section{Related Work}
\label{sec:related}

In this section, we present an overview of research pertaining to decentralised contribution estimation from three distinct viewpoints. Firstly, we review the current literature on data sharing in supply chains. Secondly, we examine the application of XAI in the industry to highlight the potential use cases and explore the extent to which value can be derived from current centralised methodologies. Thirdly, we survey the existing literature on uncertainty estimation in machine learning, which serves as the foundation for our decentralised approach.

\subsection{Data Sharing in Supply Chains}
\label{Dep}

Data sharing is an often prescribed solution to improve visibility and analytics in supply chains driven by factors such as the management of deviations and decision-making support as studied by \cite{ijpr_barriers}. Their Delphi study determined a mean importance value of 6.1 and 5.76 out of 7 for managing deviations and decision-making support. These two main factors are addressable with contribution estimation methods as shown in Fig. \ref{fig:central} and \ref{fig:decentral} by providing insights on sources of variation and allowing for pinpointed investigation of these sources to uncover and resolve problems \cite{senoner2022using}. \cite{ijpr_barriers} reports that experts state the unwillingness to share data as one of the main barriers, which is a critical problem, as the willingness to share data (6.29 mean importance out of 7) is also the second highest enabling factor listed after data quality (6.38 mean importance out of 7) by experts.

A typical SCM example is the bullwhip effect, which is characterised by amplified fluctuations in inventory and production caused by small changes in consumer demand. These fluctuations could be reduced by sharing information amongst SC actors but barriers, as discussed in \cite{ijpr_barriers}, such as trust, confidential data, and integration into existing IT systems hinder implementation. The fear of losing information advantage as described in \cite{ijpr_barriers} is further shown in a study from \cite{dai2022two}. They investigated a dual-channel supply chain with a manufacturer and retailer that can share heterogeneous information about demand uncertainty. Their results show that depending on the competitive context, one of the participants can be hurt economically. This risk further complicates the real-life adoption of data sharing. To address such problems, \cite{bechtsis2022data} recently proposed a generalised data-sharing framework with three categories of data privacy and data monetisation options. In the first category, confidential data is stored locally and not shared. The second category is confidential data that is non-critical and can be shared across the SC ecosystem. The third category is public data that can be shared with third parties and customers.
By limiting the data that is shared to non-critical information, the problem highlighted in \cite{dai2022two} could be partially addressed. Only sharing public data can avoid creating an information disadvantage for one party but at the same time, it is unlikely that most or all relevant information is contained within this non-confidential data, making this approach ineffective. For problems such as the detection of the contributing factors to a metric, we need access to process-level data which would be classified as highly critical. Hence, the framework proposed by \cite{bechtsis2022data} is not adequate to address this problem.

Confidential data can be made accessible with centralised and decentralised privacy-preserving machine learning approaches with varying levels of accuracy loss (e.g., due to added random noise when using differential privacy) \cite{boulemtafes2020review}. Given that most cross-company processes in the industry have features that are heterogeneous (e.g., production parameters of machines), the popular decentralised method of federated learning becomes increasingly hard to implement and train. Centralised methods that rely on encryption provide a method that does not reveal the underlying data without compromising data quality compared to methods such as differential privacy. These methods are challenged by the hesitance of companies to host their data in an external location and the associated technical overhead of setting up such a system.

\subsection{XAI in Industry}
\label{XAI}

With the increasing availability of data and compute resources in the age of Industry 4.0, AI applications in the industry become more prevalent. Due to the value at stake in industrial AI applications, XAI not only becomes increasingly important to ensure buy-in from stakeholders but also to understand the driving factors of outcomes in complex processes \cite{kotriwala2021xai} \cite{ahmed2022artificial}. \cite{ahmed2022artificial} conducted a survey on AI and XAI applications in Industry 4.0. Table \ref{tab:XAI} summarises the publications surveyed by \cite{ahmed2022artificial} that were classified in the XAI area. Notably, the main XAI method used is SHAP \cite{lundberg2017unified} in four of seven papers with the shared objective of determining the features that have a significant influence on the outcome of the model. We, therefore, use SHAP as the benchmark in the following chapters of this paper. The remaining three papers use Local Interpretable Model Agnostic Explanation (LIME) \cite{ribeiro2016should} and bespoke XAI methods. The use cases vary from detecting the causes of quality variance to understanding why customers decided to buy a product but still share the same common question of "Who or what contributes to the (un)desired outcome?". This application of XAI deviates from the usual XAI task of understanding why a machine learning black box model makes a decision. The mentioned papers use a machine learning model to replicate the behaviour of an industry process. By applying XAI to such a model, they can then add XAI to understand the underlying process better, essentially using the machine learning model as a surrogate for the real process. 

\begin{table}
\begin{tabular}{|c|p{170pt}|c|}
\hline
\textbf{Author} & \textbf{Main Contribution}                                              & \textbf{SHAP} \\ \hline
\cite{gramegna2020buy}                   & Determination of customer buy and leave decision factors                & \checkmark                        \\ \hline
\cite{serradilla2020interpreting}                   & Explanation of factors for residual life estimations                    &                         \\ \hline
\cite{senoner2022using}                      & Decision model for process quality improvement                          & \checkmark                      \\ \hline
\cite{meister2021investigations}               & Fault identification in composite manufacturing                         & \checkmark                      \\ \hline
\cite{mehdiyev2021explainable}                   & Post-hoc explanation method for monitoring problems                     &                       \\ \hline
\cite{brito2022explainable}                   & Fault diagnosis and anomaly detection methods in rotating machinery     & \checkmark                      \\ \hline
\cite{kharal2020explainable}                   & Fault analysis and penetration harvesting in steel plates manufacturing &                      \\ \hline
\end{tabular}
\caption{Overview of XAI-based methods in Industry 4.0 adapted from \cite{ahmed2022artificial}}
\label{tab:XAI}
\end{table}

As highlighted in Table \ref{tab:XAI}, SHAP values are one of the most popular tools for XAI in the industry due to their performance and wide availability for different models ranging from trees to deep neural networks. Based on cooperative game theory, SHAP calculates the contribution of each feature to the model prediction \cite{lundberg2017unified}.
Insights on feature contribution can be used in the industry to detect the sources of quality variance as performed by \cite{senoner2022using}, who used SHAP values to detect the processes to prioritise for quality improvement and were able to reduce yield loss by 21.7\%.

However, as soon as all relevant data is no longer accessible in one location (e.g., due to data privacy), the discussed approaches are no longer possible and novel approaches are needed to allow for contribution estimation in settings without data sharing as shown in Fig. \ref{fig:decentral}.

\subsection{Uncertainty Estimation}
\label{uncertainty}
Even though deep neural networks achieve impressive results across a wide range of tasks, the estimation of predictive uncertainty remains a challenging problem. One possible solution are Bayesian neural networks but they are non-trivial to train compared to their non-Bayesian counterparts. \cite{lakshminarayanan2017simple} therefore designed a neural network ensemble approach that is easier to train and achieves similar performance to Bayesian neural networks. This approach allows for the parallelisation of the ensembles during training and requires little hyperparameter tuning, making it ideal for industrial applications where quick deployment with little tuning is required. This is especially important for SMEs which often lack sophisticated AI capabilities. Each ensemble predicts two outputs: The label of the original problem that a single output neural network would also predict $\mu(x)$, and the variance $\sigma^2(x)$ of the prediction. By changing the loss function to minimising the negative log-likelihood (NLL) for an assumed heteroscedastic Gaussian distribution and randomly initialising the parameters of each neural network, results close to a Bayesian neural network can be achieved. The authors also propose further improvements by adding adversarial examples to the training process. \cite{lakshminarayanan2017simple} uses the NLL to evaluate the predictive uncertainty of the model output. The mean and variance of the ensemble outputs are aggregated under the assumption that the ensemble prediction is a Gaussian. 

Successive works of other authors have built upon this ensemble approach to close the gap to real Bayesian inference \cite{pearce2020uncertainty} and to expand to other popular machine learning methods such as gradient-boosted trees \cite{malinin2020uncertainty}. \\

Our research aims to combine the three fields of data sharing in supply chains, XAI in industry, and uncertainty estimation to close the gap of data privacy preserving contribution estimation in supply chains.

\section{Methodology}
\label{sec:method}

In this section, we describe the data used for the empirical validation of our method and introduce the developed framework for decentralised contribution estimation.

\subsection{Data}

\begin{figure}
\centerline{\includegraphics[width=\columnwidth]{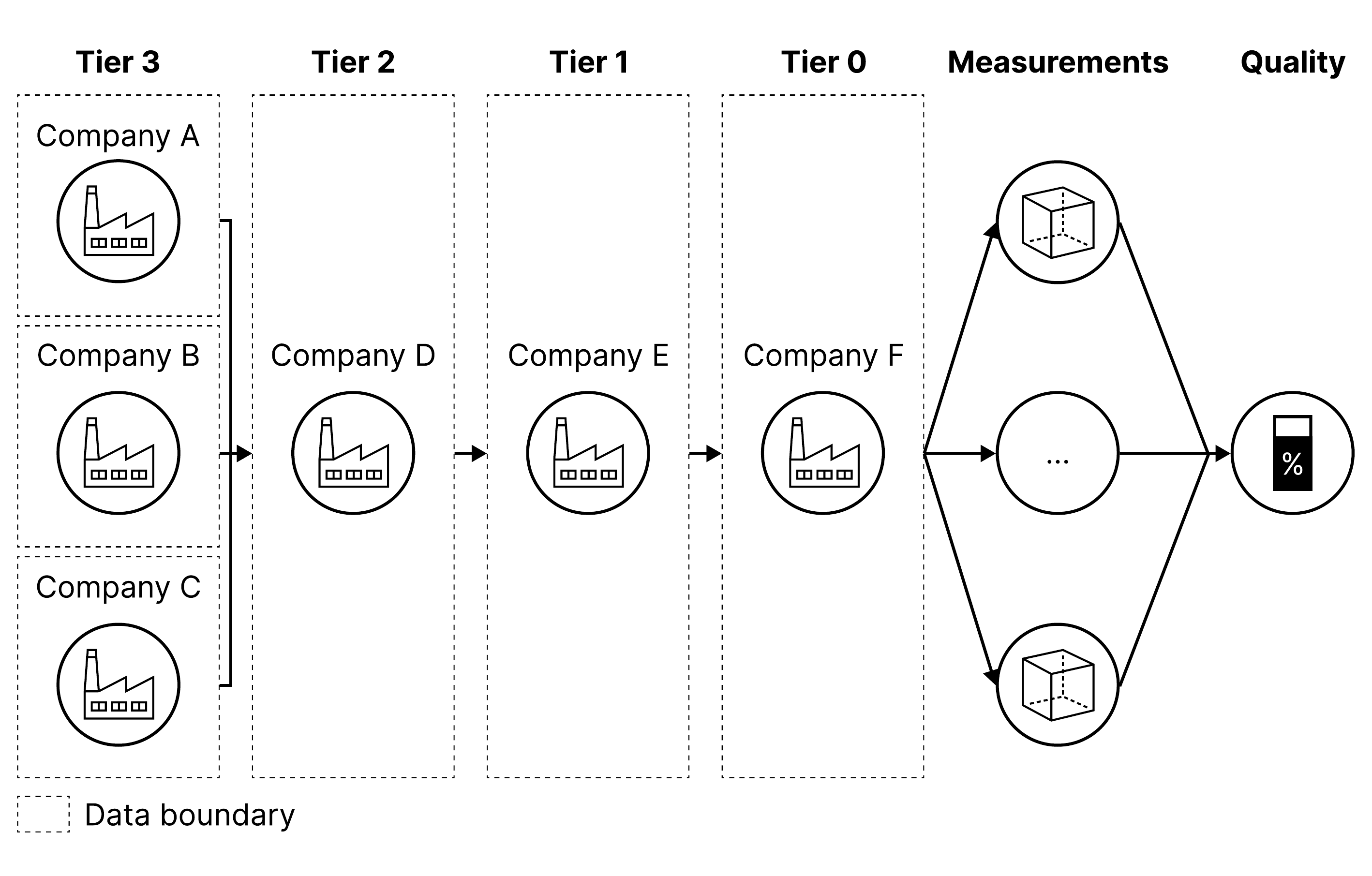}}
\caption{Multi-stage process flow of production system with measurement aggregation into a quality score}
\label{fig:process}
\end{figure}

Real-life data from multi-stage processes across companies are not publicly available due to data privacy concerns. Therefore, we use the multi-stage production data from \cite{Multista51:online} that was captured in 2019 at a production line in Detroit, Michigan by Liveline Technologies and has been used in numerous publications such as \cite{akinsolu2023generalized}, \cite{oleghe2020predictive}, \cite{wu2022synchronous}, \cite{zhang2021path}, and \cite{zhou2021attention}.
We can view each of the process steps shown in Fig. \ref{fig:process} as an individual company separated from the rest in terms of data sharing. 

\begin{figure}
\centerline{\includegraphics[width=\columnwidth]{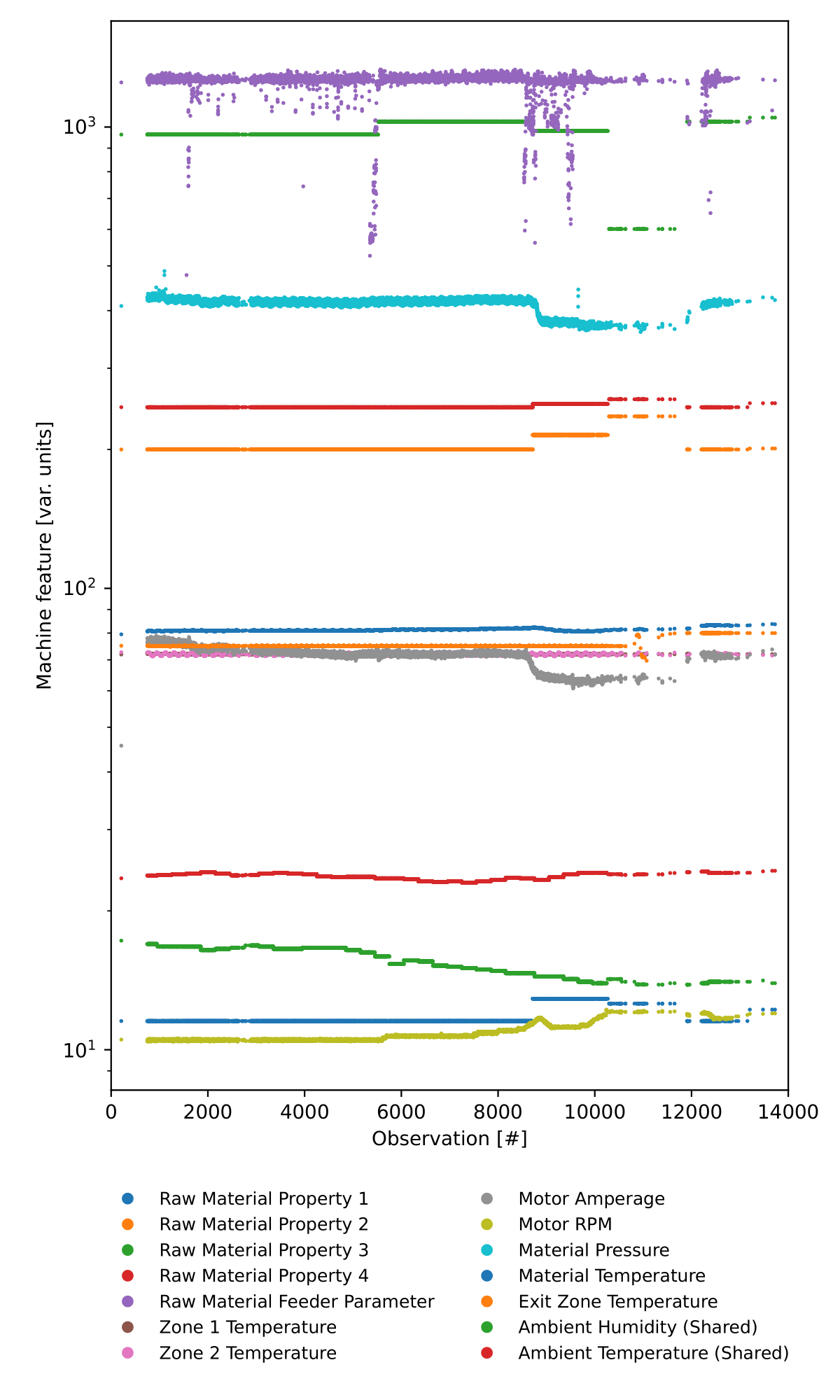}}
\caption{Feature values for Company A with datapoints for measuring errors removed}
\label{fig:features}
\end{figure}

Tier 3 suppliers in the process act in parallel and their outputs are combined by the Tier 2 process step before being passed on to Tier 1 and the original Tier 0 respectively. The outputs from the combined processes are then measured. We aggregate these measurements into one quality value to reflect a realistic Key Performance Indicator (KPI) for monitoring of this production process.
Each of the defined companies has multiple process features associated with it such as machine tool rotations per minute, raw material properties, temperatures, and more. All feature values are continuous as shown in Fig. \ref{fig:features} for Company A. The companies in the supply chain share two features that are observable to all of them. These are the ambient humidity and temperature during the manufacturing process. Features that are observable by multiple parties in real-life supply chains can range from weather data to public financial information and are therefore important to include in our case study example.

The data gaps in the observations in Fig. \ref{fig:features} are due to observations with faulty measurements, as shown in Table \ref{tab:missing}, that we dropped from the data to avoid the introduction of bias due to imputation for the target value. Measurement 4 was removed due to 12.780 of 14.000 values missing which likely indicates a faulty measurement device or process.

\begin{table}
\centering
\begin{tabular}{|c|c|}
\hline
\textbf{Measurement} & \textbf{Missing Values} \\ \hline
0                    & 1.227                   \\ \hline
1                    & 680                     \\ \hline
2                    & 608                     \\ \hline
3                    & 776                     \\ \hline
4                    & 12.780                  \\ \hline
5                    & 192                     \\ \hline
6                    & 191                     \\ \hline
7                    & 349                     \\ \hline
8                    & 977                     \\ \hline
9                    & 4.211                   \\ \hline
10                   & 638                     \\ \hline
11                   & 631                     \\ \hline
12                   & 259                     \\ \hline
13                   & 196                     \\ \hline
14                   & 917                     \\ \hline
\end{tabular}
\caption{Missing values in measurements out of 14.000 observations}
\label{tab:missing}
\end{table}

\begin{figure}
\centerline{\includegraphics[width=\columnwidth]{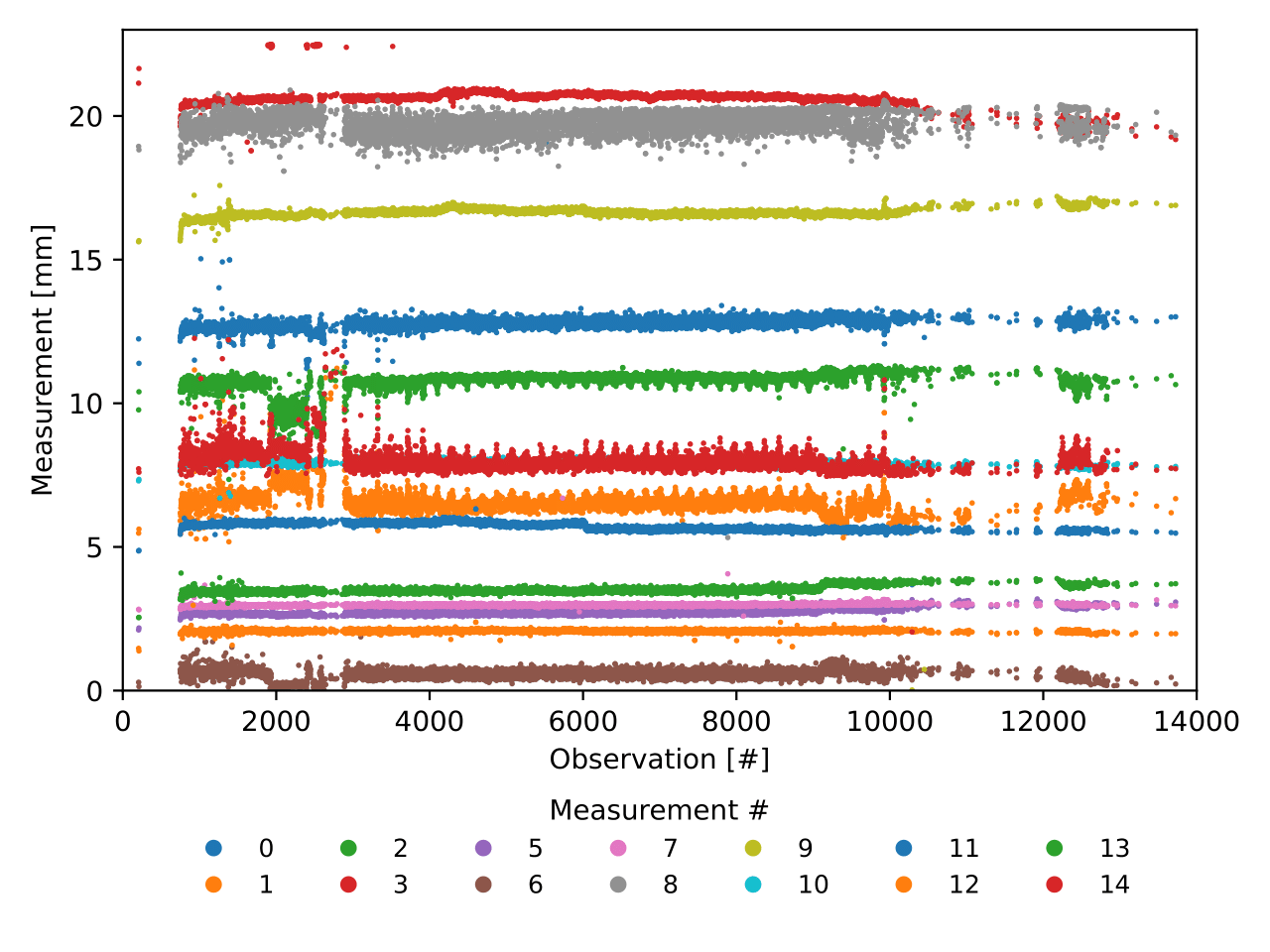}}
\caption{Measurements with datapoints for measuring errors removed (excl. \#4 due to \>90\% of data missing)}
\label{fig:measurements}
\end{figure}

The measurements without the missing value observations are displayed in Fig. \ref{fig:measurements}. The data gaps indicate disturbances in the measuring process at the beginning and towards the end of production. We aggregate these measurements into a single-quality KPI via equation \ref{eq:quality} where $n_m$ is the number of different measurements to be taken per part and $n$ the number of observations for which one measurement per $n_m$ is taken. 

\begin{equation}
\Delta Q = \frac{1}{n_m} * \sum_{i=0}^{n} \frac{|m_i-m_{i_{avg}}|}{m_{i_{avg}}}
\label{eq:quality}
\end{equation}

Due to a lack of knowledge about the measurements (e.g., is the setpoint an upper bound for functionality) we use a simple absolute relative average of the measurements.

\begin{figure}
\centerline{\includegraphics[width=\columnwidth]{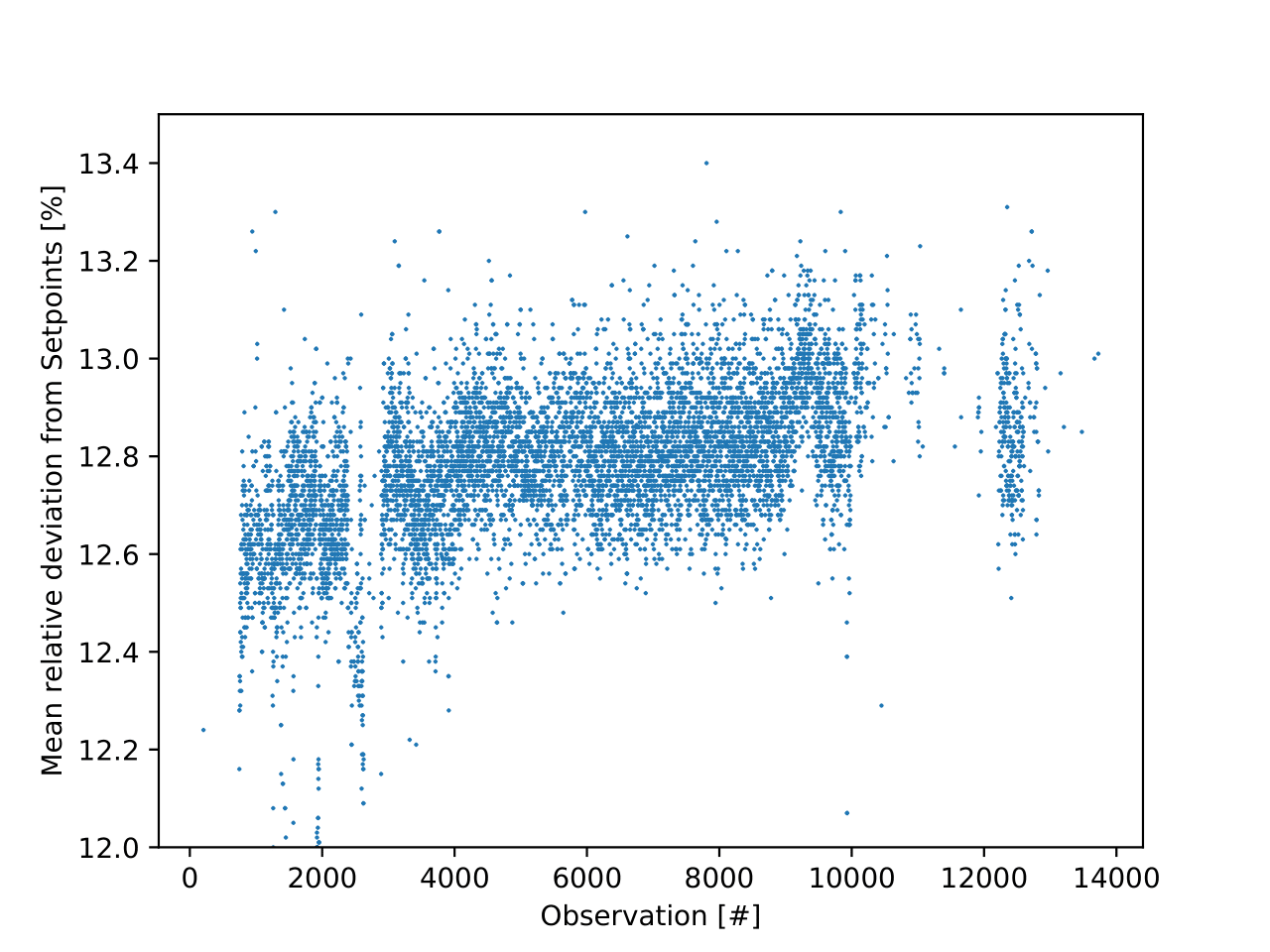}}
\caption{Aggregation of measurements into a single quality KPI with datapoints for measuring errors removed}
\label{fig:quality}
\end{figure}

The aggregation of the measurements from Fig. \ref{fig:features} via equation \ref{eq:quality} results in the quality KPI shown in Fig. \ref{fig:quality}. Due to leaving a buffer away from the specified measurements in manufacturing and our use of the absolute values, the quality KPI does not centre around 0. As we are interested in identifying the sources that impact the KPI overall, this offset does not impact our experiment.

\subsection{Decentralised contribution estimation framework}
\label{sec:methoddetail}

The problem of estimating the contribution of individual actors to the outcome of the process they are involved in can be split into two subchapters. First, the communication between actors and the associated flow of data ensures sufficient data privacy and scalability of the approach. Second, the decentralised computations can be combined and compared to determine the main contributors.

\begin{figure}
\centerline{\includegraphics[width=\columnwidth]{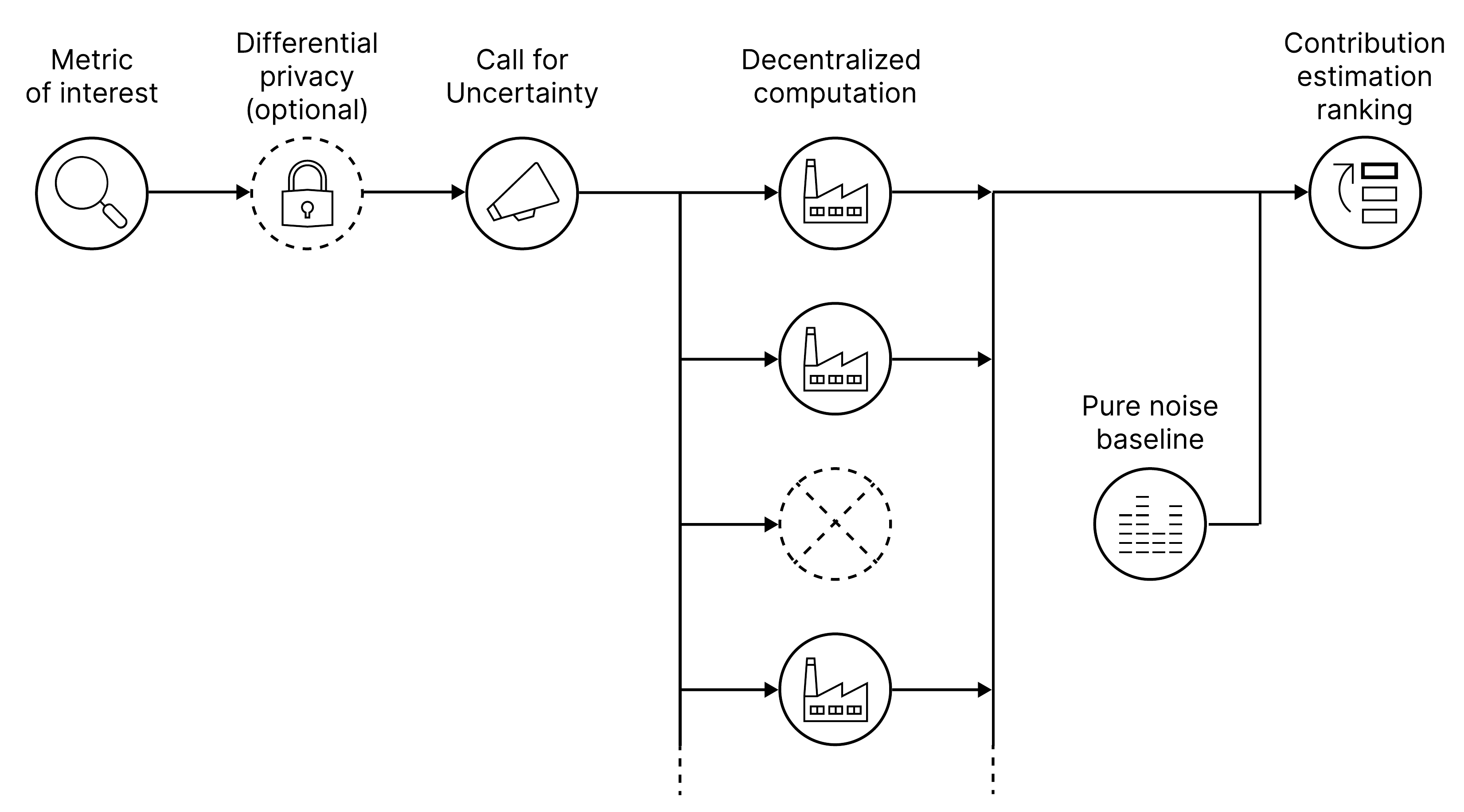}}
\caption{Decentralised contribution estimation process from selection of metric to contribution estimation with an exemplary denial of the call from one company}
\label{fig:flow}
\end{figure}

A high-level flowchart demonstrating our approach is shown in Fig. \ref{fig:flow}. Given our data from the manufacturing plant, our metric of interest is the quality of the products. This is the metric for which we want to identify the key contributors. If we wanted to enhance the privacy of the quality data to be shared, we could either rescale and offset the values or even use approaches such as differential privacy. We would then share our metric of interest with a call for uncertainty to the actors in our overall process. They can then perform the local computations of their total uncertainty score. By only sharing this single value back with us, they do not need to disclose any internal process data. If an actor does not partake, as indicated with the crossed-out circle in Fig. \ref{fig:flow} the approach still works. We perform the same uncertainty estimation as with the actors in the process using pure noise to get a baseline for what no knowledge means to identify actors that do not have a significant influence on our metric of interest. We then combine all uncertainty values including the noise-based value to create a ranking. The uncertainty in increasing order is our contribution estimation.

\subsection{Communication and data flow}

The data sharing within our proposed approach is limited to two types. First, the metric of interest for which we want to know who contributes how much. Second, the total uncertainty in predicting this metric per actor involved in the overall process.

To be able to match the metric of interest (i.e., quality of a product) to the right features within each actor, we need to also share an identifier with the metric (e.g., part ID that is traceable throughout the supply chain). This results in an array with one column of an identifier and the metric of interest. As shown in Fig. \ref{fig:flow}, the metric of interest can be modified for added privacy. This can be a modification that does not impact data quality such as rescaling or shifting the data, or methods such as differential privacy, which would greatly increase privacy by adding noise but would also add noise to the following computations.

The actors that choose to respond to the call for uncertainty reply with a single value, representing their total uncertainty. This way no internal data leaves the actors that respond to the call except for one high-level value. The only actor that shares internal data at a more granular level is the actor that wants to improve this metric and thus has the highest incentive to share anyway.

After determining the main contributors, the actor that led the call for uncertainty can start to negotiate with the main contributors on changes to improve the metric of interest and what kind of compensation they would require for this change. 

\subsection{Computation of contributions}

As described in \ref{uncertainty}, using an ensemble of neural networks to estimate the uncertainty of a prediction instead of Bayesian neural networks achieves approximately Bayesian results with greatly simplified training procedures \cite{lakshminarayanan2017simple}. We, therefore, favour this approach for industrial application. 

\begin{figure}[h]
\centerline{\includegraphics[width=\columnwidth]{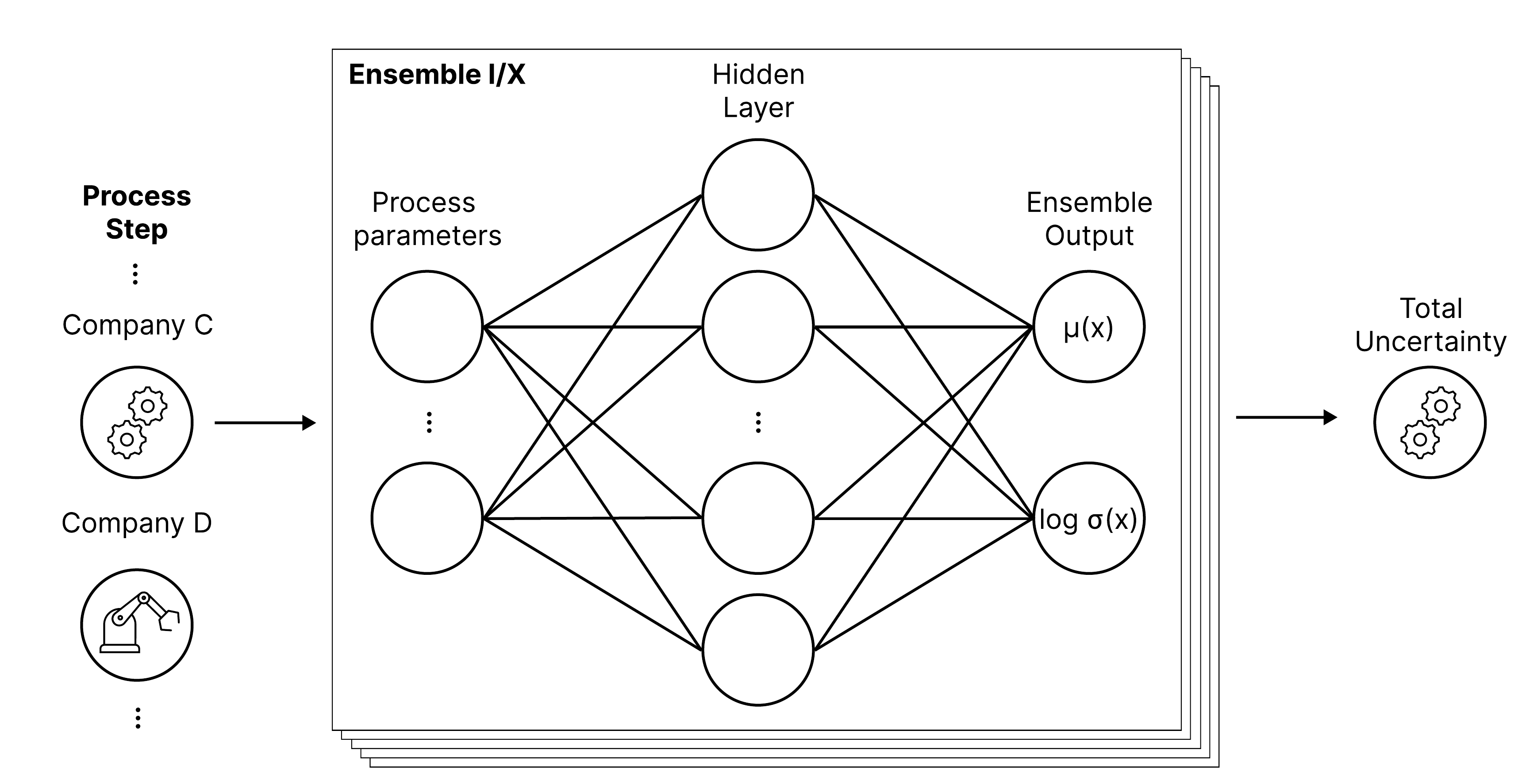}}
\caption{Decentralised ensemble approach for total uncertainty calculation per contributing actor}
\label{fig:ensemble}
\end{figure}

Our design thus follows the work of \cite{lakshminarayanan2017simple} with the parameters listed in Table \ref{tab:param} and uses early stopping with training patience of 100 or 200 epochs depending on the dataset (i.e., ending training if the model does not improve within the set number of epochs to avoid overfitting to the data). As shown in Fig. \ref{fig:ensemble}, we expand the ensemble approach depicted in the rectangular box by applying it to multiple process steps in parallel as depicted in Fig. \ref{fig:flow}. The uncertainty estimation per process step is then used to create a ranking of contribution estimation.
Based on the finding of \cite{lakshminarayanan2017simple}, we randomly initialise the weights and biases of the neural networks instead of subsampling data to increase the training data per ensemble for better results.

\begin{table}[t!]
\centering
\begin{tabular}{|l|c|}
\hline
\textbf{Parameter}   & \textbf{Value} \\ \hline
Number of ensembles $M$  & 10              \\ \hline
Size of hidden layer & 50           \\ \hline
Activation function & ReLU           \\ \hline
Dropout              & 0.5            \\ \hline
Batch size           & 128            \\ \hline
\end{tabular}
\caption{Ensemble parameter values}
\label{tab:param}
\end{table}

Each ensemble with the parameters $\{\theta_m\}{M \atop m=1}$ outputs two values, the prediction $\mu(x)$ (e.g., quality) and the log of the standard deviation of the prediction $\log(\sigma(x))$. Assuming a (heteroscedastic) Gaussian distribution, we then minimize the negative log-likelihood, as shown in equation \ref{eq:loss}, during training to include the optimisation of uncertainty estimation \cite{lakshminarayanan2017simple}.

\begin{equation}
    -\log p_\theta(y|x)= \frac{\log(\sigma^2_\theta(x))}{2}+\frac{(y-\mu_\theta(x))^2}{2*\sigma_\theta^2(x)} + \text{constant}.
    \label{eq:loss}
\end{equation}

For the quantification of uncertainty, we deviate from \cite{lakshminarayanan2017simple} where the negative log-likelihood is used. We derive the uncertainty estimates via the law of total variation which can be computed as shown in equation \ref{eq:total} \cite{malinin2020uncertainty}. The first term of the equation corresponds to the knowledge uncertainty (epistemic uncertainty) and the second term to the expected data uncertainty (aleatoric uncertainty). Together, they sum up to the total uncertainty.

\begin{equation}
\begin{split}
    \mathbb{V}_{\mathrm{p}(y \mid \boldsymbol{x}, \mathcal{D})}[y] \approx
    \frac{1}{M} \sum_{m=1}^M\left[\left(\sum_{m=1}^M \frac{\mu_m}{M}\right)-\mu_m\right]^2 \\ + \frac{1}{M} \sum_{m=1}^M \sigma_m^2 \quad,\left\{\mu_m, \sigma_m\right\}=f\left(\boldsymbol{x} ; \boldsymbol{\theta}^{(m)}\right).
    \label{eq:total}
\end{split}
\end{equation}

By using this approach, we fully decouple the total uncertainty value that will be reported from the prediction label. Equation \ref{eq:total} only contains model information but no prediction labels in contrast to the negative log-likelihood shown in equation \ref{eq:loss} used by \cite{lakshminarayanan2017simple} to quantify uncertainty. This adds additional privacy to lower the hesitation of actors to participate.
To determine which ensembles actually contain information beyond noise, we also fit one set of ensembles to random noise as shown in Fig. \ref{fig:flow}. We then rank all total uncertainty values in ascending order to get an estimation of contribution to the metric of interest.

\section{Experimental Results}
\label{sec:results}

To test the performance of our decentralised approach, we set a baseline with a centralised approach. We combine the data from all companies shown in Fig. \ref{fig:process} and train a neural network to predict the quality of the final output. The centralised model is then used to compute the SHAP values aggregated at the company level (i.e., the combination of all feature importances per company). These are used as the ground truth for comparison of the contribution estimation.

\begin{figure}[t!]
\centerline{\includegraphics[width=\columnwidth]{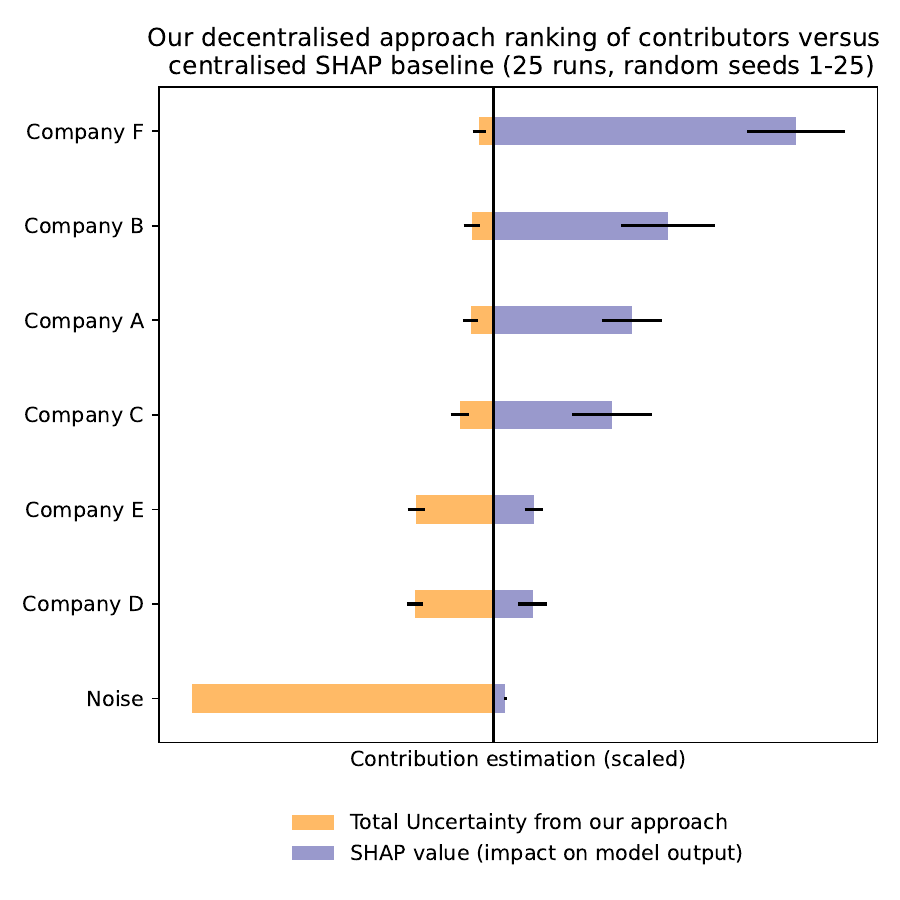}}
\caption{Multi-stage process flow: Contribution estimation comparison of our decentralised approach with a centralised SHAP approach}
\label{fig:results_detroit}
\end{figure}

The results of our decentralised approach compared to the centralised SHAP approach are shown in Fig. \ref{fig:results_detroit}. For ease of reading, we have scaled the minimum and maximum values of SHAP and our uncertainty prediction to be equal and plotted the uncertainty prediction in ascending order. Our underlying reasoning is, that higher uncertainty indicates less knowledge about the outcome and thus less influence of the process parameters on the metric of interest. As highlighted in Fig. \ref{fig:flow}, we add one prediction with pure noise to create a baseline for what "no contribution" means. As expected, the pure noise prediction has the highest uncertainty and thus implies the lowest influence on our metric of interest. All result figures report mean and standard deviations for 25 different seed values to show the robustness of the approach.

Within Fig. \ref{fig:results_detroit}, we can observe that the ranking holds true for all actors. Our decentralised approach correctly identifies company F as the main contributor in line with the results from the centralised SHAP approach. The results for companies A, B, and C also correctly represent the findings from the centralised SHAP approach. It should be noted that both the SHAP approach and the uncertainty approach have significant overlaps in their prediction ranges given the reported standard deviations for companies A, B, and C but are clearly separated from companies E and D, as well as noise.

\begin{figure}[t!]
\centerline{\includegraphics[width=\columnwidth]{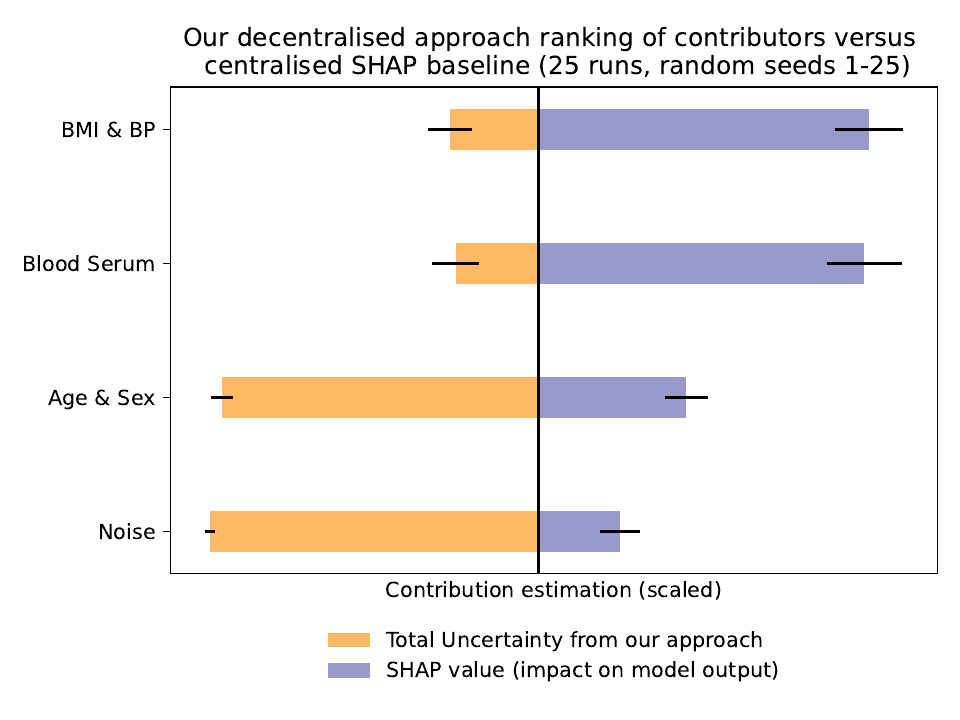}}
\caption{Diabetes risk: Contribution estimation comparison of our decentralised approach with a centralised SHAP approach}
\label{fig:diabetes}
\end{figure}

\begin{figure}[t!]
\centerline{\includegraphics[width=\columnwidth]{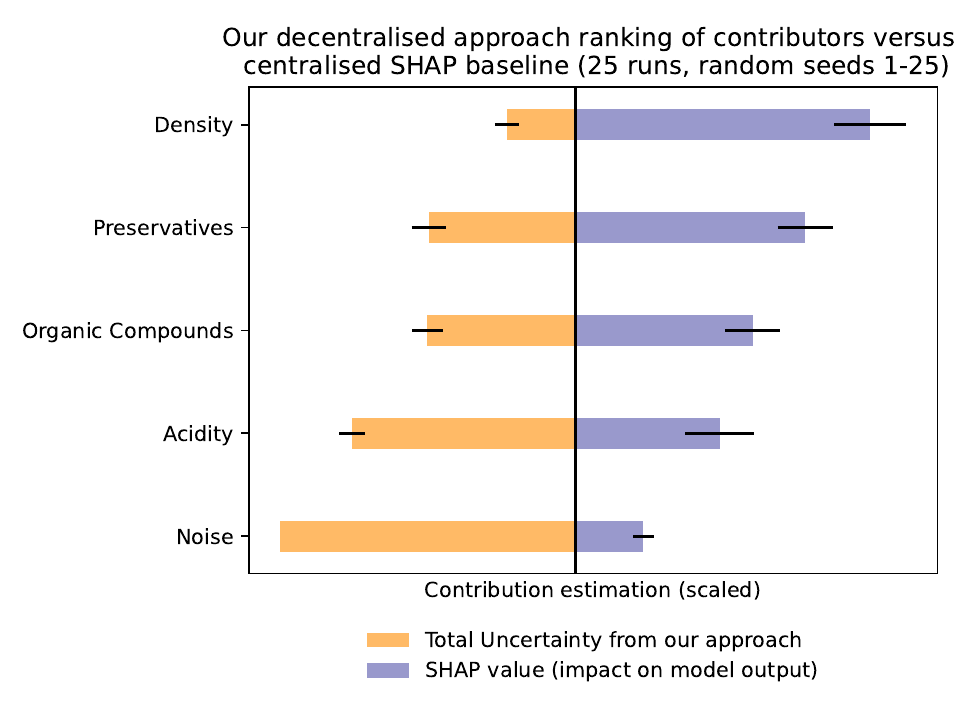}}
\caption{Red wine quality: Contribution estimation comparison of our decentralised approach with a centralised SHAP approach}
\label{fig:red_wine}
\end{figure}

\begin{figure}[h!]
\centerline{\includegraphics[width=\columnwidth]{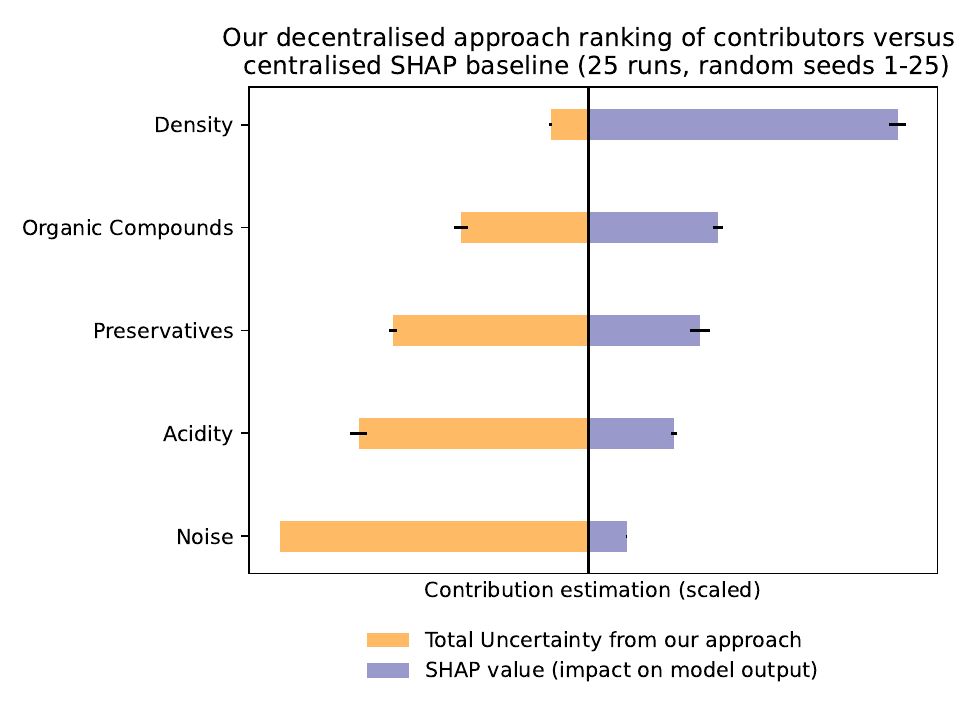}}
\caption{White wine quality: Contribution estimation comparison of our decentralised approach with a centralised SHAP approach}
\label{fig:white_wine}
\end{figure}

In order to further verify our method on non-sequential problems, we use three popular tabular data regression problems from \cite{Lichman:2013}: Diabetes, red wine quality, and white wine quality. We combine similar features together into logical groups (e.g., organic compounds) and treat them as independent data sources similar to the supply chain problem. The results are shown in Fig. \ref{fig:diabetes}, \ref{fig:red_wine}, and \ref{fig:white_wine} again showing comparable results to the centralised SHAP approach while preserving data privacy. By using the same neural network architecture across all experiments, we further show the generalisability of the approach.

Our new decentralised contribution estimation method provides decision-makers with previously unavailable insights that are comparable to what they would be able to achieve with a centralised approach.

\section{Conclusion}
\label{sec:conclusion}

This work proposes a decentralised contribution estimation method to address the
main shortcoming of XAI methods in SC, data privacy. We demonstrate the viability of our decentralised approach by comparing it to a centralised SHAP approach in a multi-echelon supply chain setting. Our approach achieves similar results while preserving data privacy and furthermore shows a clearer differentiation between pure noise data and information-containing data.
The main limitation of our approach is, that we do not account for multicollinearity between features in different companies.
In practice, our method enables companies to efficiently identify the main contributors to a metric of interest. They can then collaborate to find a win-win situation in which one actor adjusts their process to improve the metric of interest in return for compensation from the other actor that can be paid out of the gained overall value of the improvement. This win-win setting also acts as an incentive for companies to participate in the process.
Our method can be further leveraged in applications such as federated learning in industry. We can use the results of our method to select the main contributors to the metric of interest (e.g., risk prediction) and then only implement the federated learning setup with those selected companies. By removing companies with low contributions to the overall outcome, costs for the setup and running of federated learning can be saved. A further practical SCM usage in line with recent AI developments are autonomous SCM agents that represent companies. Our method can be used to help these agents to determine who to interact with in order to resolve problems such as delivery time fluctuations. The key benefit of our method in this setting is the ability to quickly and safely determine the key agents to interact with out of all supply chain participants, drastically reducing the problem size.

\bibliographystyle{IEEEtran}
\bibliography{ref}

\begin{thebibliography}{10}
\providecommand{\url}[1]{#1}
\csname url@samestyle\endcsname
\providecommand{\newblock}{\relax}
\providecommand{\bibinfo}[2]{#2}
\providecommand{\BIBentrySTDinterwordspacing}{\spaceskip=0pt\relax}
\providecommand{\BIBentryALTinterwordstretchfactor}{4}
\providecommand{\BIBentryALTinterwordspacing}{\spaceskip=\fontdimen2\font plus
\BIBentryALTinterwordstretchfactor\fontdimen3\font minus \fontdimen4\font\relax}
\providecommand{\BIBforeignlanguage}[2]{{%
\expandafter\ifx\csname l@#1\endcsname\relax
\typeout{** WARNING: IEEEtran.bst: No hyphenation pattern has been}%
\typeout{** loaded for the language `#1'. Using the pattern for}%
\typeout{** the default language instead.}%
\else
\language=\csname l@#1\endcsname
\fi
#2}}
\providecommand{\BIBdecl}{\relax}
\BIBdecl

\bibitem{zantek2002process}
P.~F. Zantek, G.~P. Wright, and R.~D. Plante, ``Process and product improvement in manufacturing systems with correlated stages,'' \emph{Management Science}, vol.~48, no.~5, pp. 591--606, 2002.

\bibitem{senoner2022using}
J.~Senoner, T.~Netland, and S.~Feuerriegel, ``Using explainable artificial intelligence to improve process quality: Evidence from semiconductor manufacturing,'' \emph{Management Science}, vol.~68, no.~8, pp. 5704--5723, 2022.

\bibitem{meister2021investigations}
S.~Meister, M.~Wermes, J.~St{\"u}ve, and R.~M. Groves, ``Investigations on explainable artificial intelligence methods for the deep learning classification of fibre layup defect in the automated composite manufacturing,'' \emph{Composites Part B: Engineering}, vol. 224, p. 109160, 2021.

\bibitem{brito2022explainable}
L.~C. Brito, G.~A. Susto, J.~N. Brito, and M.~A. Duarte, ``An explainable artificial intelligence approach for unsupervised fault detection and diagnosis in rotating machinery,'' \emph{Mechanical Systems and Signal Processing}, vol. 163, p. 108105, 2022.

\bibitem{new_barbarossa2007decentralized}
S.~Barbarossa and G.~Scutari, ``Decentralized maximum-likelihood estimation for sensor networks composed of nonlinearly coupled dynamical systems,'' \emph{IEEE Transactions on Signal Processing}, vol.~55, no.~7, pp. 3456--3470, 2007.

\bibitem{new_jiang2021differential}
B.~Jiang, J.~Li, G.~Yue, and H.~Song, ``Differential privacy for industrial internet of things: Opportunities, applications, and challenges,'' \emph{IEEE Internet of Things Journal}, vol.~8, no.~13, pp. 10\,430--10\,451, 2021.

\bibitem{new_shang2023robust}
S.~Shang, X.~Li, K.~Gu, L.~Li, X.~Zhang, and V.~Pandi, ``A robust privacy-preserving data aggregation scheme for edge-supported iiot,'' \emph{IEEE Transactions on Industrial Informatics}, 2023.

\bibitem{new_ying2023privacy}
C.~Ying, N.~Zheng, Y.~Wu, M.~Xu, and W.-A. Zhang, ``Privacy-preserving adaptive resilient consensus for multi-agent systems under cyber attacks,'' \emph{IEEE Transactions on Industrial Informatics}, 2023.

\bibitem{new_tableak_fl}
M.~Vero, M.~Balunovi\'{c}, D.~I. Dimitrov, and M.~Vechev, ``Tableak: tabular data leakage in federated learning,'' in \emph{Proceedings of the 40th International Conference on Machine Learning}, ser. ICML'23.\hskip 1em plus 0.5em minus 0.4em\relax JMLR.org, 2023.

\bibitem{new_FL1}
\BIBentryALTinterwordspacing
X.~Xu, J.~Wu, M.~Yang, T.~Luo, X.~Duan, W.~Li, Y.~Wu, and B.~Wu, ``Information leakage by model weights on federated learning,'' in \emph{Proceedings of the 2020 Workshop on Privacy-Preserving Machine Learning in Practice}, ser. PPMLP'20.\hskip 1em plus 0.5em minus 0.4em\relax New York, NY, USA: Association for Computing Machinery, 2020, p. 31–36. [Online]. Available: \url{https://doi.org/10.1145/3411501.3419423}
\BIBentrySTDinterwordspacing

\bibitem{lakshminarayanan2017simple}
B.~Lakshminarayanan, A.~Pritzel, and C.~Blundell, ``Simple and scalable predictive uncertainty estimation using deep ensembles,'' \emph{Advances in neural information processing systems}, vol.~30, 2017.

\bibitem{lundberg2017unified}
S.~M. Lundberg and S.-I. Lee, ``A unified approach to interpreting model predictions,'' \emph{Advances in neural information processing systems}, vol.~30, 2017.

\bibitem{ijpr_barriers}
\BIBentryALTinterwordspacing
J.~O. Tarun Kumar~Agrawal, Ravi~Kalaiarasan and M.~Wiktorsson, ``Supply chain visibility: A delphi study on managerial perspectives and priorities,'' \emph{International Journal of Production Research}, vol.~62, no.~8, pp. 2927--2942, 2024. [Online]. Available: \url{https://doi.org/10.1080/00207543.2022.2098873}
\BIBentrySTDinterwordspacing

\bibitem{dai2022two}
Y.~Dai, L.~Dou, H.~Song, L.~Zhou, and H.~Li, ``Two-way information sharing of uncertain demand forecasts in a dual-channel supply chain,'' \emph{Computers \& Industrial Engineering}, vol. 169, p. 108162, 2022.

\bibitem{bechtsis2022data}
D.~Bechtsis, N.~Tsolakis, E.~Iakovou, and D.~Vlachos, ``Data-driven secure, resilient and sustainable supply chains: gaps, opportunities, and a new generalised data sharing and data monetisation framework,'' \emph{International Journal of Production Research}, vol.~60, no.~14, pp. 4397--4417, 2022.

\bibitem{boulemtafes2020review}
A.~Boulemtafes, A.~Derhab, and Y.~Challal, ``A review of privacy-preserving techniques for deep learning,'' \emph{Neurocomputing}, vol. 384, pp. 21--45, 2020.

\bibitem{kotriwala2021xai}
A.~Kotriwala, B.~Kl{\"o}pper, M.~Dix, G.~Gopalakrishnan, D.~Ziobro, and A.~Potschka, ``Xai for operations in the process industry-applications, theses, and research directions.'' in \emph{AAAI Spring Symposium: Combining Machine Learning with Knowledge Engineering}, 2021.

\bibitem{ahmed2022artificial}
I.~Ahmed, G.~Jeon, and F.~Piccialli, ``From artificial intelligence to explainable artificial intelligence in industry 4.0: a survey on what, how, and where,'' \emph{IEEE Transactions on Industrial Informatics}, vol.~18, no.~8, pp. 5031--5042, 2022.

\bibitem{ribeiro2016should}
M.~T. Ribeiro, S.~Singh, and C.~Guestrin, ``" why should i trust you?" explaining the predictions of any classifier,'' in \emph{Proceedings of the 22nd ACM SIGKDD international conference on knowledge discovery and data mining}, 2016, pp. 1135--1144.

\bibitem{gramegna2020buy}
A.~Gramegna and P.~Giudici, ``Why to buy insurance? an explainable artificial intelligence approach,'' \emph{Risks}, vol.~8, no.~4, p. 137, 2020.

\bibitem{serradilla2020interpreting}
O.~Serradilla, E.~Zugasti, C.~Cernuda, A.~Aranburu, J.~R. de~Okariz, and U.~Zurutuza, ``Interpreting remaining useful life estimations combining explainable artificial intelligence and domain knowledge in industrial machinery,'' in \emph{2020 IEEE International Conference on Fuzzy Systems (FUZZ-IEEE)}.\hskip 1em plus 0.5em minus 0.4em\relax IEEE, 2020, pp. 1--8.

\bibitem{mehdiyev2021explainable}
N.~Mehdiyev and P.~Fettke, ``Explainable artificial intelligence for process mining: A general overview and application of a novel local explanation approach for predictive process monitoring,'' \emph{Interpretable Artificial Intelligence: A Perspective of Granular Computing}, pp. 1--28, 2021.

\bibitem{kharal2020explainable}
A.~Kharal, ``Explainable artificial intelligence based fault diagnosis and insight harvesting for steel plates manufacturing,'' \emph{arXiv preprint arXiv:2008.04448}, 2020.

\bibitem{pearce2020uncertainty}
T.~Pearce, F.~Leibfried, and A.~Brintrup, ``Uncertainty in neural networks: Approximately bayesian ensembling,'' in \emph{International conference on artificial intelligence and statistics}.\hskip 1em plus 0.5em minus 0.4em\relax PMLR, 2020, pp. 234--244.

\bibitem{malinin2020uncertainty}
A.~Malinin, L.~Prokhorenkova, and A.~Ustimenko, ``Uncertainty in gradient boosting via ensembles,'' \emph{arXiv preprint arXiv:2006.10562}, 2020.

\bibitem{Multista51:online}
``Multi-stage continuous-flow manufacturing process | kaggle,'' \url{https://www.kaggle.com/datasets/supergus/multistage-continuousflow-manufacturing-process}, (Accessed on 03/30/2023).

\bibitem{akinsolu2023generalized}
M.~O. Akinsolu and K.~Zribi, ``A generalized framework for adopting regression-based predictive modeling in manufacturing environments,'' \emph{Inventions}, vol.~8, no.~1, p.~32, 2023.

\bibitem{oleghe2020predictive}
O.~Oleghe, ``A predictive noise correction methodology for manufacturing process datasets,'' \emph{Journal of Big Data}, vol.~7, no.~1, pp. 1--27, 2020.

\bibitem{wu2022synchronous}
Z.~Wu, Y.~Li, and L.~Hu, ``A synchronous multiple change-point detecting method for manufacturing process,'' \emph{Computers \& Industrial Engineering}, vol. 169, p. 108114, 2022.

\bibitem{zhang2021path}
D.~Zhang, Z.~Liu, W.~Jia, H.~Liu, and J.~Tan, ``Path enhanced bidirectional graph attention network for quality prediction in multistage manufacturing process,'' \emph{IEEE Transactions on Industrial Informatics}, vol.~18, no.~2, pp. 1018--1027, 2021.

\bibitem{zhou2021attention}
X.~Zhou and X.~Gao, ``An attention-based forecasting network for intelligent services in manufacturing,'' in \emph{Service-Oriented Computing: 19th International Conference, ICSOC 2021, Virtual Event, November 22--25, 2021, Proceedings 19}.\hskip 1em plus 0.5em minus 0.4em\relax Springer, 2021, pp. 900--914.

\bibitem{Lichman:2013}
\BIBentryALTinterwordspacing
M.~Lichman, ``{UCI} machine learning repository,'' 2013. [Online]. Available: \url{http://archive.ics.uci.edu/ml}
\BIBentrySTDinterwordspacing

\end{thebibliography}

\end{document}